\documentclass[journal]{IEEEtran}
\usepackage{times}
\usepackage{latexsym}
\usepackage{graphicx}
\usepackage{amsmath}
\usepackage{amsthm}
\usepackage{booktabs}
\usepackage{algorithm}
\usepackage{algorithmic}
\usepackage{amssymb}
\usepackage{multirow}
\usepackage{multicol}
\usepackage{tabularx}
\usepackage{arydshln}
\usepackage{etoolbox}
\usepackage{url}
\usepackage{ulem}
\makeatletter
\patchcmd{\@makecaption}
  {\scshape}
  {}
  {}
  {}
\makeatletter
\patchcmd{\@makecaption}
  {\\}
  {.\ }
  {}
  {}
\makeatother
\def\tablename{Table}

\ifCLASSINFOpdf
\else
\fi
\begin{document}
\title{Multi-grained Evidence Inference for Multi-choice Reading Comprehension}

\author{Yilin~Zhao, Hai~Zhao, Sufeng~Duan
\IEEEcompsocitemizethanks{
\IEEEcompsocthanksitem{This paper was partially supported by Joint Research Project of Yangtze River Delta Science and Technology Innovation Community (No. 2022CSJGG1400).}
\IEEEcompsocthanksitem Yilin~Zhao, Hai~Zhao and Sufeng~Duan are with the Department of Computer Science and Engineering, Shanghai Jiao Tong University, and also with Key Laboratory of Shanghai Education Commission for Intelligent Interaction and Cognitive Engineering, Shanghai Jiao Tong University. Yilin Zhao and Sufeng Duan contributed equally to this work. Corresponding author: Hai Zhao.
\protect\\ E-mail: zhaoyilin@sjtu.edu.cn, zhaohai@cs.sjtu.edu.cn, 1140339019dsf@sjtu.edu.cn.}}

\markboth{IEEE/ACM TRANSACTIONS ON AUDIO, SPEECH, AND LANGUAGE PROCESSING}%
{Shell \MakeLowercase{\textit{Zhao et al.}}: Multi-grained Evidence Inference for Multi-choice Reading Comprehension}

\normalem
\newcommand{\tabincell}[2]{\begin{tabular}{@{}#1@{}}#2\end{tabular}}
\maketitle

\begin{abstract}
Multi-choice Machine Reading Comprehension (MRC) is a major and challenging task for machines to answer questions according to provided options.
Answers in multi-choice MRC cannot be directly extracted in the given passages, and essentially require machines capable of reasoning from accurate extracted evidence. 
However, the critical evidence may be as simple as just one word or phrase, while it is hidden in the given redundant, noisy passage with multiple linguistic hierarchies from phrase, fragment, sentence until the entire passage. 
We thus propose a novel general-purpose model enhancement which integrates multi-grained evidence comprehensively, named \emph{\textbf{Mu}lti-\textbf{g}rained \textbf{e}vidence i\textbf{n}ferencer (\textbf{Mugen})}, to make up for the inability.
\emph{Mugen} extracts three different granularities of evidence: coarse-, middle- and fine-grained evidence, and integrates evidence with the original passages, achieving significant and consistent performance improvement on four multi-choice MRC benchmarks.
\end{abstract}


%
\IEEEpeerreviewmaketitle

\section{Introduction}
As a fundamental and challenging task of natural language understanding (NLU), Machine Reading Comprehension (MRC) requires machines to answer questions according to the given passages \cite{zhang2020survey}.
According to the differences in expectant answers, MRC tasks can be divided into three common formats \cite{khashabi2020unifiedqa,baradaran2020survey}: 1) extractive task, which searches for the most proper snippet from the passage as answer \cite{rajpurkar2016squad,yang2018hotpotqa}; 
2) generative task, which needs model to summarize the passage and generate answer \cite{Schwarz2018narrativeqa};
3) multi-choice task, the focus of this work, which provides several options and aims to select the most suitable one \cite{richardson2013mctest,lai2017race}.

Though multi-choice MRC seems not so challenging that the answers have been shown among candidate options, the real difficulty is, the answers together with their supported evidence may not appear explicitly in the given passages at all.
Thus to perform the multi-choice MRC satisfactorily, there comes an essential demand requiring the models capable of inference based on accurate and abundant evidence.

However, questions in multi-choice MRC may accompany with lengthy passages with noise, which hide critical evidence in different levels:

1) Evidence may appear in a quite refined level, and in some cases, one phrase or even one word can determine the prediction of the question;

2) Evidence may hide in quite diverse grained units inside the passage, which needs to infer from the information among phrases, fragments, sentences, until the entire passage.

\begin{figure*}[ht]
\centering
\includegraphics[scale=1.0]{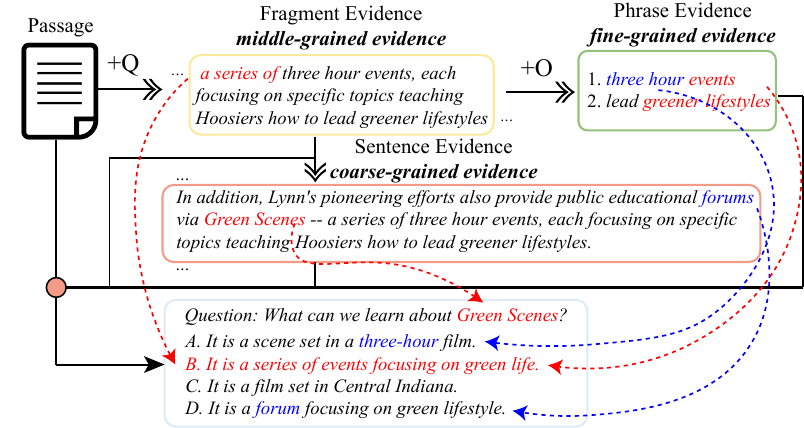}
\caption{Sample process of \emph{Mugen}. ``+Q” and ``+O” respectively represent \emph{Mugen} extracts refined evidence based on question and options. Red/blue lines respectively represent golden/interference evidence chains.}
\label{example}
\end{figure*}

One example from RACE \cite{lai2017race} is shown as Figure \ref{example}.
To answer the given question, extraction and integration of the complete golden evidence chain (marked in red) distributed in different linguistic levels are necessary.
Relying on each single level may lead to incomprehensive explanation and inference (for \emph{fine-grained evidence}), or introduce interference information which leads to incorrect prediction (marked in blue, mostly for \emph{coarse-grained evidence}).

Though well-extracted evidence rather than the entire passage for later inference is essential to solve concerned MRC tasks effectively, most existing studies only obtain single-grained evidence in a rough way \cite{wang2019evidence} and fail to make a flexible and comprehensive multi-grained evidence processing, leading to marginal improvements.
Inspired by raising studies with ``coarse-to-fine” and ``multi-grained” thoughts \cite{choi2017coarse,zheng2020multi-grained}, we propose a concise model which pays attention to the evidence in multiple granularities, called \emph{\textbf{Mu}lti-\textbf{g}rained \textbf{e}vidence i\textbf{n}ferencer (\textbf{Mugen})}.
As Figure \ref{example} shows, \emph{Mugen} first extracts \emph{middle-grained evidence} in a fragment level, then finds out the sentences containing it as \emph{coarse-grained evidence}, as well as extracts a set of critical phrases as \emph{fine-grained evidence}.
With the integration of the original passage and three different granularities of evidence, \emph{Mugen} products the evidence-enhanced prediction.
The effectiveness of \emph{Mugen} is verified on four multi-choice MRC benchmarks: RACE, DREAM, Cosmos QA and MCTest, and obtains substantial performance improvement over strong baselines by passing MRC significance tests \cite{zhang2020retro}.

\section{Related Studies}
In recent years, more challenging MRC tasks among various forms have been proposed \cite{wang2018glue,rajpurkar2018SQuAD2,huang2019CosmosQA}.
To solve MRC tasks, researchers train powerful pre-trained models and obtain significant improvements \cite{devlin2019bert,liu2019roberta,yang2019xlnet,lewis2020bart}.
With the raising encoding ability of pre-trained contextual encoders, some researchers try to quote external commonsense \cite{lin2019kagnet,shwartz2020unsupervised} or train on additional profitable datasets \cite{khashabi2020unifiedqa} to enhance their models in an outer way.

In the meantime, more researchers attempt to strengthen models in an inner way without external information.
Some studies improve the interaction embedding between input sequences based on attention networks \cite{zhang2020dcmn,wang2018co,tang2019multi}, while other studies focus on human reading strategy simulation \cite{li2018unified,sun2020context,zhao2022poinet}.
Among the inductive strategies, \emph{evidence extraction} plays an important role \cite{sun2019strategy,niu2020soft-evidence}.

However, most existing studies are limited to the evidence in one single granularity, which may reduce the attention on critical phrase information (for coarse-grained evidence, like \cite{choi2017coarse,wang2019evidence}) or lack complete contextual explanation (for fine-grained evidence, like \cite{ma2020xtqa}).

Raising studies with ``coarse-to-fine” or ``multi-grained” thoughts for \textbf{non-MRC} tasks provide a possible solution for the above limitations.
For open-domain QA, Zhong \cite{zhong2019coarse} and Zheng \cite{zheng2020multi-grained} utilize multi-grained co-attentions to encode documents and score answers.
And for long document extractive QA, Choi \cite{choi2017coarse} use coarse-to-fine reading strategies for single-grained evidence evaluation.
However, no previous work applies the above ``multi-grained” thought to MRC field especially challenging multi-choice MRC.

Inspired by the previous works of evidence enhancement and multi-grained strategy, this paper proposes \textit{Mugen} to make the first attempt to integrate multi-grained evidence comprehensively for inference enhancement in the MRC field, and achieves inspiring results with concise design, highlighting the effectiveness of hierarchical evidence extraction and integration.

\section{Our Model}
We focus on multi-choice MRC in this work, which can be represented as a triple $<P, Q, O>$, where $P$ is a passage, $Q$ is a question over $P$, $O=\{O_1, O_2, ... O_U\}$ is a set of options for $Q$, and $U$ is the number of options.
Among the options, the most appropriate option $O_{gold}$ has been chosen as the ground truth answer, and the goal of our model is to pick up the answer $O_{gold}$.
Thus we let the model learn subject to:
$$\mathbb{P}(O_{gold}\mid P, Q, O)\ge \mathbb{P}(O_i\mid P, Q, O), \quad i \in \{1,2,...,U\},$$
where $\mathbb{P}$ represents probability.

\subsection{Multi-grained Evidence}
In this work, three different grains of evidence are proposed for multi-grained evidence integration and modeling enhancement, where:
\begin{itemize}
    \item As \emph{coarse-grained} evidence, \textbf{Sentence Evidence (Set)} is one single sentence (or a set of sentences) that contains the critical evidence in the lengthy passage, with appropriate rich contextual information.
    \item As \emph{middle-grained} evidence, \textbf{Fragment Evidence} is the shortest sub-sentence fragment with complete linguistic structures\footnote{As the middle-grained flexible granularity, the typical case of Fragment Evidence is a clause sentence, but it can convert from several phrases to nearly the entire sentence.}.
    Fragment Evidence is used to extract the most concise and explicit text segment with complete semantics as evidence, for answer prediction in the subsequent processes.
    Thus in most cases, we can find Fragment Evidence possesses good interpretability, like the examples in Table \ref{evidence}.
    \item As \emph{fine-grained} evidence, \textbf{Phrase Evidence Set} is a set of “feature” phrases in the middle-grained evidence.
    Different from Fragment Evidence, most phrases in the Phrase Evidence only have adequate complete meanings, rather than complete linguistic structures.
    Therefore, the main function of Phrase Evidence is to further highlight critical words or phrases, rather than serve as interpretable evidence texts directly.
\end{itemize}

\begin{table*}[ht]
\centering
\caption{\label{evidence} Evidence of different granularity in sample documents and conversations. The corpora are from RACE and DREAM.}
\begin{tabular}{|p{17.5cm}|}
\hline \bf Document 1\\ \hline
\emph{... Also known as the Scarce Tortoiseshell, it has an orange and blue colour and is about one third bigger than our own Small Tortoiseshell. Butterfly Conservation was starting its annual Big Butterfly Count,a yearly survey of the butterflies across the nation. Sir David Attenborough, President of the charity, said, the UK is a nation of amateur naturalists and we have a proud tradition of celebrating and studying our wildlife. ...}\\\hdashline
\textbf{Q}: \emph{The annual Big Butterfly Count is intended to \_ .}\quad\quad\textbf{A}:\emph{study butterflies across Britain}\\\hline
\textbf{Sentence Evidence}: \textit{Butterfly Conservation was starting its annual Big Butterfly Count, a yearly survey of the butterflies across the nation.}\\
\textbf{Fragment Evidence}: \textit{a yearly survey of the butterflies across the nation}\\
\textbf{Phrase Evidence}: \textit{a yearly survey; the butterflies across the nation}\\\hline\hline \bf Document 2\\ \hline
\emph{... Supporters of online relationships state that the Internet allows couples to get to know each other intellectually first. Personal appearance doesn't get in the way. But critics of online relationships argue that no one can truly know another person in cyberspace. ...}\\\hdashline
\textbf{Q}: \emph{People who are against online love think \_ .}\quad\quad\quad\quad\textbf{A}:\emph{one may not show the real self in cyberspace}\\\hline
\textbf{Sentence Evidence}: \textit{But critics of online relationships argue that no one can truly know another person in cyberspace.}\\
\textbf{Fragment Evidence}: \textit{no one can truly know another person in cyberspace}\\
\textbf{Phrase Evidence}: \textit{truly know another person; cyberspace}\\\hline\hline \bf Conversation 1\\ \hline
\emph{...}\\
\emph{A: Why did you choose to be an author?}\\
\emph{B: Well, if you want to achieve happiness, step one would be finding out what you love doing most. Step two would be finding someone to pay you to do this. I consider myself very lucky to be able to support myself by writing.}\\
\emph{...}\\\hdashline
\textbf{Q}: \emph{Why does Ms. Rowling consider herself so lucky?}\quad\quad\textbf{A}:\emph{She can make a living by writing.}\\\hline
\textbf{Sentence Evidence}: \textit{I consider myself very lucky to be able to support myself by writing.}\\
\textbf{Fragment Evidence}: \textit{be able to support myself by writing}\\
\textbf{Phrase Evidence}: \textit{support myself; writing}\\\hline
\end{tabular}
\end{table*}

Table \ref{evidence} shows several samples of multi-grained evidence in both textual and conversational corpora, where the evidence in each granularity has a relatively complete semantic and syntactic structure, and provides critical information for answer prediction.

\subsection{Overall Framework}
\begin{figure*}[ht]
\centering
\includegraphics[scale=0.7]{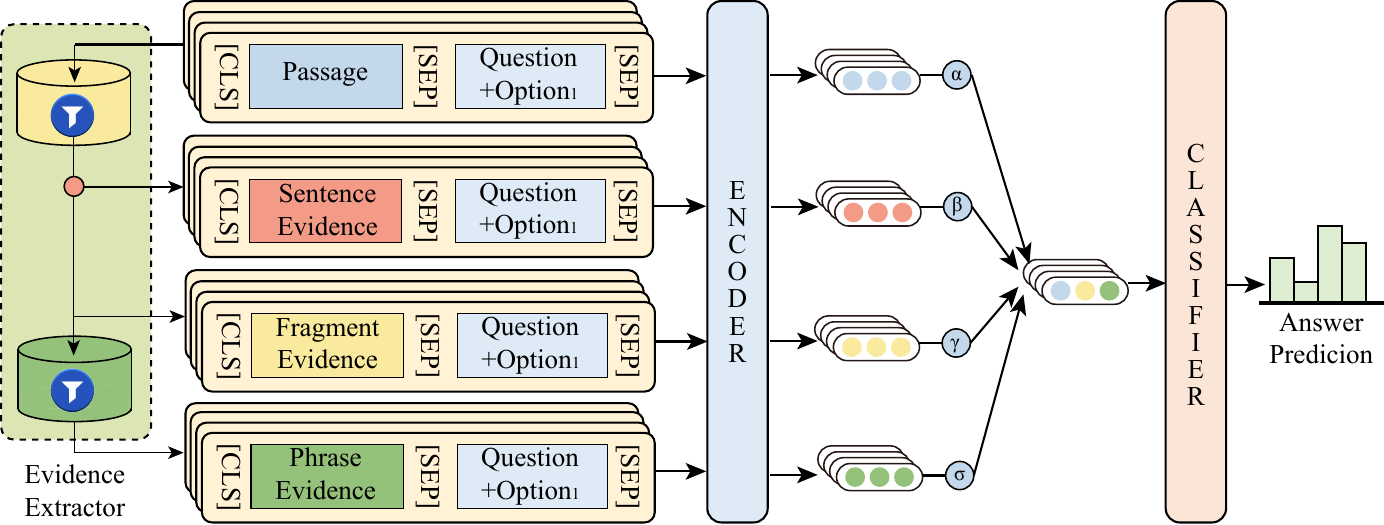}
\caption{The overview of \emph{Mugen}. $\alpha, \beta, \gamma$ and $\sigma$ are integrating weight coefficients, satisfying $\alpha + \beta + \gamma + \sigma = 1.0$.}
\label{model}
\end{figure*}

The overall framework of \emph{Mugen} is shown in Figure \ref{model}.
With the help of \emph{Evidence Extractor}, \emph{Mugen} filters out Sentence Evidence, Fragment Evidence and Phrase Evidence respectively, as the \emph{coarse-}, \emph{middle-} and \emph{fine-grained evidence}.
If the evidence set in a certain granularity contains more than one textual piece, \emph{Mugen} will splice these pieces by space.

Then \emph{Mugen} encodes the above evidence with the question and options respectively, and executes a weighted integration of them for prediction.
In detail, \emph{Mugen} uses its baseline as the \emph{Encoder} (a single parameter-sharing ALBERT \cite{lan2019albert} in this work) to encode the textual content of the evidence in each granularity, as well as the complete contextual
information of the passage.

In the separate encoding process, as the granularity of encoded evidence becomes finer, the input sequence of the encoder contains less contextual information.
As a result, contextual information takes up less proportion in the embedding representation of the finer-grained evidence, while the textual content of critical evidence takes up more.

The subsequent integration process can be formulated as:
$$E^i = dropout (\alpha e_{pas}^i + \beta e_{sen}^i + \gamma e_{fra}^i + \sigma e_{phr}^i) \in \mathbb{R}^H,$$
where $H$ is the hidden size of \emph{Encoder}, and  $\alpha,\beta,\gamma,\sigma$ are learnable parameters.
$E^i, e_{pas}^i, e_{sen}^i, e_{fra}^i$ and $e_{phr}^i$ represent the $[CLS]$ embedding vector from the last hidden layer of ``Question + $i$-th Option” with the final evidence-enhanced representation, the original passage, sentence evidence, fragment evidence and phrase evidence respectively.

In the above process, with the integration of the passage embedding ($e_{pas}$) and evidence embeddings ($e_{sen}/e_{fra}/e_{phr}$), \textit{Mugen} integrates the contextual representation of the entire passage and the enhanced textual representation of each single-grained evidence.
Thus among the information embedded in the evidence-enhanced embedding $E$, critical evidence occupies a greater proportion, leading to a more accurate answer prediction.

In \emph{Mugen}, a softmax layer as the \textit{Classifier} is employed to calculate scores for options, and the total loss is the standard Cross Entropy Loss between the integrated prediction and the golden answer:
$$\mathcal{L}=-\frac{1}{U}\sum_{i=1}^U (Bool(O_i=O_{gold})\cdot log(p_i))),$$
$$p_i=\frac{\exp(w_i^TE_i+b_i)}{\sum_{j=1}^U \exp(w_j^TE_j+b_j)},$$
where $w_i \in \mathbb{R}^H, b_i \in \mathbb{R}^1$ are learnable parameters.

\subsection{Evidence Extractor}
As Figure \ref{model} shows, there are two sub-extractors in the \emph{Evidence Extractor}: \emph{Sentence Evidence Extractor} (the upper one) and \emph{Phrase Evidence Extractor} (the lower one).

\noindent$\bullet$ \textbf{Sentence Evidence Extractor}\\
\emph{Mugen} uses \emph{Sentence Evidence Extractor} to extract both Sentence and Fragment Evidence.
To ensure \emph{Sentence Evidence Extractor} can extract precise evidence, we implement a contextual encoder (we employ ALBERT$_{base}$ \cite{lan2019albert} in \emph{Mugen}) which is pre-trained on SQuAD 2.0 \cite{rajpurkar2018SQuAD2} individually to extract a non-null answer span\footnote{We eliminate the possibility of extracting null spans by drastically increasing the threshold $\tau$ in the above encoder. According to \cite{devlin2019bert}, when $S \cdot T_0+E \cdot T_0 > \max_{i\leq j} S \cdot T_i+E \cdot T_j + \tau$, the encoder will extract a null span, where $T_i \in \mathbb{R}^H$ is the embedding of the $i$-th input token, and $S/E\in \mathbb{R}^H$ is the introduced start/end vector.}.
Then the extracted span is defined as the Fragment Evidence for \emph{Mugen}.
Benefiting from the pre-training on SQuAD 2.0, \emph{Sentence Evidence Extractor} can ensure the segmenting correctness and linguistic integrity of Fragment Evidence to a large extent.

In addition, though \emph{Sentence Evidence Extractor} can be modified to extract several pieces of Fragment Evidence, we only retain one piece with the highest confidence score, because the benchmarks we focus on do not have obvious multi-hop features like MultiRC \cite{Khashabi2018MultiRC}.
Multiple weak-relevant fragments may reduce the proportion of critical information in the entire Fragment Evidence, causing inference deviation with further extraction and integration.

In the next step, \textit{Mugen} obtains Sentence Evidence Set based on Fragment Evidence.
If Fragment Evidence locates in one single sentence $S$, then $S$ is the only element in Sentence Evidence Set; and if Fragment Evidence spans several consequent sentences $\{S_1, ..., S_k\}$, then sentences $\{S_1, ..., S_k\}$ are added into the Sentence Evidence Set\footnote{In most cases, Fragment Evidence is the subsection of one single sentence.}.

\begin{figure}[ht]
\centering
\includegraphics[scale=0.85]{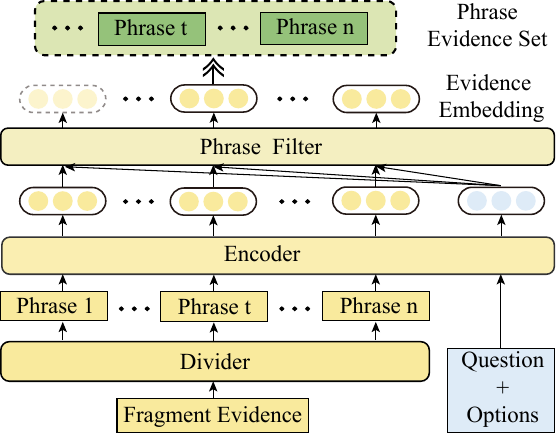}
\caption{The overview of \emph{Phrase Evidence Extractor}.}
\label{phrase}
\vspace{-0.3cm}
\end{figure}

\noindent$\bullet$ \textbf{Phrase Evidence Extractor}\\
Based on Fragment Evidence, \emph{Phrase Evidence Extractor} extracts Phrase Evidence as fine-grained evidence, shown in Figure \ref{phrase}.

In \textit{Divider}, the Fragment Evidence is divided into $n$ phrases: $\{Phrase_1,...,Phrase_n\}$ based on stopwords (including prepositions, pronouns, conjunctions and interjections)\footnote{Some prepositions (like ``from”) are retained because they can express specific meanings in some specific phrases (such as ``come from”).} and punctuation.
\emph{Mugen} removes the above words and punctuation, and splits the fragments before and after them into independent phrases.
With minor computational cost, the above rule-based phrase segmentation method highlights critical words and phrases in the Fragment Evidence, and makes the phrases get appropriate segmentation in most cases.

After that, \emph{Mugen} splices the question with all given options with spaces, encodes the above question and phrases by an ALBERT$_{base}$ Encoder, and calculates the correlation scores of their embedding vectors:
$$s_i=p_i^Tq, i\in (1,...,n),$$
where $s_i$ and $p_i$ are respectively the correlation score and embedding of $Phrase_i$, $q$ is the embedding of the question with options.

In \emph{Phrase Filter}, \emph{Mugen} retains all $Phrase_i$  satisfying: $s_i > \theta \times s_{max}$, where $s_{max}$ is the maximum correlation score among all phrases, and $\theta$ is the evidence threshold.
Finally, all retained phrases form the Phrase Evidence Set.
By splicing these phrases with spaces, \emph{Mugen} generates the ultimate Phrase Evidence.

\subsection{Simplified Version of Mugen}
In the above \emph{Mugen}, multiple runs of the baseline encoder are required to integrate and determine an appropriate proportion of multi-grained evidence, which may ask for higher computational cost.
To control such extra computational cost, we simplify the multi-grained evidence integration method in \textit{Mugen}, providing \textit{Mugen$_{simp}$}.
There is only one input sequence of the processed passage in \textit{Mugen$_{simp}$}, therefore it only requires a single run of encoding, without additional computational cost beyond the baseline model.
In the processed passage text, there are $6$ special tags ($<sos><eos><sof><eof><sop><eop>$) around the evidence in $3$ different granularities, and each granularity has $2$ tags labeling its start and end positions.
For example, Table \ref{example_tag} shows the input sample of the example passage in Figure \ref{example}, pre-processed by \textit{Mugen$_{simp}$}.

\begin{table}[ht]
\caption{\label{example_tag} Input sample of the example passage in Figure \ref{example}, pre-processed by \textit{Mugen$_{simp}$}.}
\centering
\begin{tabular}{|p{8cm}|}
\hline \bf Original Passage \\ \hline
\emph{... Indiana Living Green is the only local publication focused on green living and sustainability. In addition, Lynn's pioneering efforts also provide public educational forums via Green Scenes -- a series of three hour events, each focusing on specific topics teaching Hoosiers how to lead greener lifestyles. She is a sought-after speaker, delivering topics such as ...}\\\hline
\bf Processed Passage \\ \hline
\emph{... Indiana Living Green is the only local publication focused on green living and sustainability. $<sos>$ In addition, Lynn's pioneering efforts also provide public educational forums via Green Scenes -- $<sof>$ a series of $<sop>$ three hour events $<eop>$, each focusing on specific topics teaching Hoosiers how to $<sop>$ lead greener lifestyles $<eop>$ $<eof>$. $<eos>$ She is a sought-after speaker, delivering topics such as ...}\\\hline
\end{tabular}
\end{table}

\section{Experiments}
\subsection{Setup}
We run the experiments on $8$ NVIDIA Tesla $V100$ GPUs.
The implementation of \emph{Mugen} is based on the PyTorch \cite{paszke2019pytorch} implementation of ALBERT$_{xxlarge}$, and the hyper-parameters of \emph{Mugen} are shown in Table \ref{hyperparam}.

\begin{table}[ht]
\caption{\label{hyperparam}The fine-tuning hyper-parameters of \emph{Mugen}. LR: learning rate, BS: batch size, TE: training epochs, SS: save steps.}
\centering{
	\begin{tabular}{c|c c c c}
	    \hline	\bf Hyperparam & LR & BS & TE & SS\\
		\hline \hline
		\bf DREAM & 1e-5 & 24 & 3 & 400\\
		\bf RACE & 1e-5 & 32 & 2 & 4000\\
		\bf Cosmos QA & 1e-5 & 32 & 3 & 2000\\
		\bf MCTest 500 & 1e-5 & 24 & 2 & 50\\
		\bf MCTest 160 & 1e-5 & 24 & 6 & 50\\\hline
	\end{tabular}
}
\end{table}

As a supplement, the warmup rate is $0.1$ for all datasets, and we set $\theta=0.8$ for Phrase Evidence Extractor\footnote{With $\theta$ changing to $0.7$, $0.9$ and $1.0$, the average score of \emph{Mugen} based on ALBERT$_{base}$ on DREAM got $0.08\%$, $0.29\%$ and $0.46\%$ reduction respectively.}.
For the length of passage and evidence in different granularities, we set $512$ for passage, $128$ for \emph{Sentence Evidence}, and $32$ for \emph{Fragment Evidence} and \emph{Phrase Evidence}.

\subsection{Dataset}
We evaluate \emph{Mugen} on four multi-choice MRC benchmarks: RACE \cite{lai2017race}, DREAM \cite{sun2019dream}, Cosmos QA \cite{huang2019CosmosQA} and MCTest \cite{richardson2013mctest}.
The detailed descriptions are shown as following:

\textbf{RACE} is a large-scale MRC task collected from English examinations, which contains nearly $100,000$ questions.
Its passages are in the form of articles, and most questions need contextual reasoning.
In RACE, the average word length of the passages is $313$, and the domains of passages are diversified. 

\textbf{DREAM} is a conversation-based multi-choice MRC task, containing more than $10,000$ questions, where the average word length of the conversations is $147$.
The challenge of the dataset is that more than $80\%$ of the questions are non-extractive and require reasoning from multi-turn conversations.

\textbf{Cosmos QA} is a large-scale MRC task, which has about $35,600$ questions and the passages are collected from people’s daily narratives.
The questions are about the causes or effects of events, which can benefit from commonsense injection as well as evidence extraction.
The passages in Cosmos QA have an average word length of $71$.

\textbf{MCTest} is a multi-choice MRC task, whose passages are from fictional stories, with an average word length of $240$.
One of the challenges is that most questions require evidence dispersing in different parts of the passage, which can benefit well from our model.

\subsection{Results}
\begin{table}[ht]
\caption{\label{dream_result} Public submissions on DREAM. The accuracy results ($\%$) in the first domain are from the leaderboard.}
\centering{
	\begin{tabular}{l|c|c}
    \hline\bf Model & \bf Dev & \bf Test \\\hline\hline
    FTLM++ \cite{sun2019dream} & 58.1 & 58.2    \\
    BERT$_{base}$ \cite{devlin2019bert} & 63.4 & 63.2  \\
    BERT$_{large}$ \cite{devlin2019bert} & 66.0 & 66.8   \\
    XLNet$_{large}$ \cite{yang2019xlnet} & --  & 72.0 \\
    RoBERTa$_{large}$ \cite{liu2019roberta} & 85.4 & 85.0  \\
    RoBERTa$_{large}$ + MMM \cite{jin2019mmm} & 88.0 & 88.9 \\
    ALBERT$_{xxlarge}$ + RekNet \cite{zhao2020reknet} & 89.8 & 89.6\\
    ALBERT$_{xxlarge}$ + Retraining \cite{ju2021retraining} & 90.2 & 90.0 \\
    ALBERT$_{xxlarge}$ + DUMA \cite{zhu2020dual} & 89.9 & 90.4 \\
    ALBERT$_{xxlarge}$ + DUMA + Multi-Task Learning & -- & 91.8 \\\hline
    ALBERT$_{base}$ (rerun) & 65.7 & 65.6 \\
    Mugen$_{simp}$ on ALBERT$_{base}$ & 68.6 & 68.3 \\
    Mugen on ALBERT$_{base}$ & 68.8 & 68.7 \\\hline
    ALBERT$_{xxlarge}$ (rerun) & 88.7 & 88.3 \\
    Mugen$_{simp}$ on ALBERT$_{xxlarge}$ & 89.6 & 89.8 \\
    Mugen on ALBERT$_{xxlarge}$ & 90.1 & 90.4 \\\hline\hline
	\end{tabular}
	}
\end{table}

\begin{table} [ht]
\caption{\label{race_result} Public submissions on RACE. The accuracy results ($\%$) in the first domain are from the leaderboard. SC denotes single choice and TL denotes transfer learning.}
\centering{
	\begin{tabular}{p{3.7cm}|c|c}
    \hline\bf Model & \bf Dev (M / H) & \bf Test (M / H)\\\hline\hline
    BERT$_{base}$ \cite{devlin2019bert} & 64.6 (-- / --) & 65.0 (71.1 / 62.3)  \\
    BERT$_{large}$ \cite{devlin2019bert} & 72.7 (76.7 / 71.0) & 72.0 (76.6 / 70.1)  \\
    XLNet$_{large}$ \cite{yang2019xlnet} & 80.1 (-- / --) & 81.8 (85.5 / 80.2)\\
    XLNet$_{large}$ + DCMN+ \cite{zhang2020dcmn} & -- (-- / --) & 82.8 (86.5 / 81.3)\\
    RoBERTa$_{large}$ \cite{liu2019roberta} & -- (-- / --) & 83.2 (86.5 / 81.8)\\
    RoBERTa$_{large}$ + MMM \cite{jin2019mmm} & -- (-- / --) & 85.0 (89.1 / 83.3)\\
    T5-11B \cite{raffel2020t5} & -- (-- / --) & 87.1 (-- / --)\\
    ALBERT$_{xxlarge}$ + RekNet \cite{zhao2020reknet} & 87.8 (91.1 / 86.4) & 87.8 (90.1 / 86.8)\\
    ALBERT$_{xxlarge}$ + DUMA \cite{zhu2020dual} & 88.1 (-- / --) & 88.0 (90.9 / 86.7)\\
    ALBERT$_{xxlarge}$ + Retraining \cite{ju2021retraining} & 88.4 (91.3 / 87.2) & 88.0 (91.2 / 86.7)\\
    T5-11B + UnifiedQA \cite{khashabi2020unifiedqa} & -- (-- / --) & 89.4 (-- / --)\\
    Megatron-BERT-3.9B \cite{shoeybi2019megatron} & -- (-- / --) & 89.5 (91.8 / 88.6)\\
    ALBERT$_{xxlarge}$ + SC + TL \cite{jiang2020single} & -- (-- / --) & 90.7 (92.8 / 89.8) \\\hline
    ALBERT$_{base}$ (rerun) & 67.9 (72.3 / 65.7) & 67.2 (72.1 / 65.2) \\
    Mugen$_{simp}$ on ALBERT$_{base}$  & 71.7 (73.3 / 68.6) & 70.6 (73.5 / 67.3)\\
    Mugen on ALBERT$_{base}$  & 72.1 (73.7 / 69.1) & 71.1 (74.0 / 68.0)\\\hline
    ALBERT$_{xxlarge}$ (rerun) & 86.6 (89.4 / 85.2) & 86.5 (89.2 / 85.4) \\
    Mugen$_{simp}$ on ALBERT$_{xxlarge}$ & 88.0 (90.8 / 86.8) & 87.8 (90.5 / 86.6)\\
    Mugen on ALBERT$_{xxlarge}$ & 88.4 (91.4 / 87.1) & 88.1 (91.2 / 87.0)\\\hline\hline
	\end{tabular}
	}
\end{table}

\begin{table}[ht]
\caption{\label{cosmos_result} Public submissions on Cosmos QA leaderboard by $Mar\;1st, 2022$, reported by accuracy ($\%$). The amount of parameters in T5-11B is nearly $50$ times more than in \textit{Mugen}. Models with $*$ inject external commonsense or corpus for data augmentation in an outer way.}
\centering{
	\begin{tabular}{l|c|c}
	\hline
		\bf Model & \bf Dev & \bf Test \\\hline\hline
		T5-11B \cite{raffel2020t5} & -- & 90.3\\
		T5-11B + UNICORN$^*$ \cite{Lourie2021unicorn} & -- & 91.8\\\hline
		BERT$_{base}$ \cite{devlin2019bert} & 66.2 & 67.1\\
		RoBERTa$_{large}$ \cite{liu2019roberta} & 81.7 & 83.5\\
		RoBERTa$_{large}$ + CEGI$^*$ \cite{liu2020cegi} & 83.8 & 83.6\\
		ALBERT$_{xxlarge}$ + GDIN$^*$ \cite{tian2020gdin} & -- & 84.5\\
		RoBERTa$_{large}$ + ALICE \cite{pereira2020alice} & 83.6 & 84.6\\
		ALBERT$_{xxlarge}$ + RekNet$^*$ \cite{zhao2020reknet} & 85.9 & 85.7\\\hline
		ALBERT$_{base}$ (rerun) & 63.1 & 63.7\\
		Mugen$_{simp}$ on ALBERT$_{base}$ & 65.0 & 65.3\\
		Mugen on ALBERT$_{base}$ & 65.6 & 65.7\\\hline
		ALBERT$_{xxlarge}$ (rerun) & 85.0 & 84.8\\
		Mugen$_{simp}$ on ALBERT$_{xxlarge}$ & \bf 86.0 & \bf 85.9\\
		Mugen on ALBERT$_{xxlarge}$ & \bf 86.4 & \bf 86.2\\\hline
	\end{tabular}
}
\end{table}

\begin{table}[ht]
\caption{\label{mctest_result} Accuracy results ($\%$) on the test set of MCTest. Results in the first domain are from \cite{zhang2020dcmn}.}
\centering{
	\begin{tabular}{l|c|c}
	\hline
		\bf Model & \bf MC160 & \bf MC500\\\hline\hline
		BERT$_{large}$ \cite{devlin2019bert} & 73.8 & 80.4\\
		XLNet$_{large}$ \cite{yang2019xlnet} & 80.6 & 83.4\\
		GPT + Strategies \cite{sun2019strategy} & 81.7 & 82.0\\
		BERT$_{large}$ + DCMN \cite{zhang2020dcmn} & 85.0 & 86.5\\
		XLNet$_{large}$ + DCMN \bf(Previous SOTA) \cite{zhang2020dcmn} & 86.2 & 86.6\\\hline
		ALBERT$_{base}$ (rerun) & 73.9 & 76.9\\
        Mugen$_{simp}$ on ALBERT$_{base}$ & 81.5 & 84.3\\
		Mugen on ALBERT$_{base}$ & 82.0 & 84.7\\\hline
		ALBERT$_{xxlarge}$ (rerun) & 88.3 & 88.0\\
        Mugen$_{simp}$ on ALBERT$_{xxlarge}$ & 90.5 & 89.6\\
		Mugen on ALBERT$_{xxlarge}$ & \bf 90.9 & \bf 90.3\\\hline
	\end{tabular}
}
\end{table}

Taking \textbf{accuracy($\%$)} as the evaluation criteria, with $5$ random seeds, our average results are shown in Tables \ref{dream_result}--\ref{mctest_result}\footnote{Due to the test set of Cosmos QA is not available for free evaluations with different random seeds, we report the results with one single seed.}.
As a supplement, the average standard deviations of the development and test results of \emph{Mugen} on ALBERT$_{xxlarge}$ are respectively $0.55, 0.23, 1.14$ and $0.77$ on DREAM, RACE, MCTest $160$ and MCTest $500$, which shows \emph{Mugen} has satisfactory stability of answer prediction.

For the performance, \emph{Mugen} outperforms the strong baselines and other powerful models on the leaderboards without any external information or additional neural networks with numerous parameters like DUMA \cite{zhu2020dual} (shown in Table \ref{parameters}).
Even so, \emph{Mugen} achieves state-of-the-art (SOTA) performance on both sub-dataset of MCTest beyond the previous SOTA model \cite{zhang2020dcmn};
and SOTA performance on Cosmos QA\footnote{\url{https://leaderboard.allenai.org/cosmosqa/submissions/public}} among models with moderate contextual encoders except for two models with huge T5, due to our limited computing resources.
Besides, \emph{Mugen} passes McNemar’s significance test\footnote{In a statistical sense, if a model passes McNemar's significance test, we can conclude the performances of the evaluated model and its baseline model have a statistically significant difference. Following the settings in previous works \cite{zhao2022poinet}, we define ``whether the answer of baseline/proposed model is correct” as the pair in McNemar's test. For example, if the answer of the proposed model is correct and the baseline is wrong, the pair is $0-1$.} \cite{mcnemar1947test} with $p < 0.01$ for all the above datasets as Zhang \cite{zhang2020retro} suggested.
It indicates that, compared to the baseline model, the performance gains from \emph{Mugen} are statistically significant.
From another point of view, existing powerful pre-trained models can gain further substantial improvements from the integration of multi-grained evidence.





As for the proportions, with five random seeds, the final learned results are $\alpha=0.46$, $\beta=0.19$, $\gamma=0.28$ and $\sigma=0.07$ on average.
In this work, \emph{Mugen} is a generalized representation enhancement method for diverse tasks and baselines without advanced auxiliary tech on specific datasets \cite{raffel2020t5,jiang2020single}, and we verify \emph{Mugen} in a standardized setting.
Even so, \emph{Mugen} obtains consistent and statistical significant improvement over strong baselines, and achieves SOTA performance on two benchmarks, pointing out the prospect of deeper exploration and integration of the information in given datasets.

\begin{table}[ht]
\caption{\label{parameters} Parameter statistics in the training process of \emph{Mugen} and baselines. Parameters in \textit{Evidence Extractor} are only used for evidence extraction instead of the training process.}
\centering{
	\begin{tabular}{l|l}
	    \hline	\bf Model & \bf Parameters\\
		\hline \hline
        ALBERT$_{xxlarge}$ & 235M\\
        ALBERT$_{xxlarge}$ + DUMA & 292M (+24.3\%)\\\hline
        ALBERT$_{xxlarge}$ (rerun)& 212.29M\\
        Mugen on ALBERT$_{xxlarge}$ & 212.29M (+0.0\%)\\\hdashline
        Mugen on ALBERT$_{xxlarge}$ + Evidence Extractor & 222.87M (+\textbf{5.0\%})\\
        \hline
	\end{tabular}
}
\end{table}

In terms of parameter scale, \emph{Mugen} has no additional parameters beyond baselines during the training process, as Table \ref{parameters} shows.
In terms of computational cost, with almost no additional computation, \textit{Mugen$_{simp}$} still obtains acceptable improvement over strong baselines, reiterating that the improvement of \emph{Mugen} comes from the integration of multi-grained evidence.
We record the training time cost of models on \textit{base/xxlarge} size on RACE, \textit{Mugen} costs $44/738$ minutes for one training epoch while \textit{Mugen$_{simp}$} costs $29/366$ minutes, saving $41.9\%$ training cost on average.
Thus, we recommend \textit{Mugen$_{simp}$} to researchers who pursue lower computational cost.

\section{Analysis}
We evaluate \emph{Mugen} on ALBERT$_{base}$ on DREAM for further analysis, and experiments on other datasets like RACE show a similar quantitative tendency.

\subsection{Ablation Studies}
\begin{table}[ht]
\caption{\label{ablation_result} The accuracy results ($\%$) of ablation studies on DREAM.}
\centering{
	\begin{tabular}{p{4cm}|c|c}
	    \hline	\bf Model & \bf Dev & \bf Test  \\
		\hline \hline
        Baseline (ALBERT$_{base}$)  & 65.74 & 65.56\\
        Ensemble Baseline & 66.87 & 66.73\\
        Mugen on ALBERT$_{base}$ & 68.83 & 68.69\\
        \quad - Sentence Evidence & 67.99 & 67.82\\
        \quad - Fragment Evidence & 67.55 & 67.66\\
        \quad - Phrase Evidence & 68.18 & 68.05\\\hline
        Baseline (ALBERT$_{xxlarge}$)  & 88.69 & 88.28\\
        Ensemble Baseline & 88.87 & 88.49\\
        Mugen on ALBERT$_{xxlarge}$ & 90.19 & 90.42\\
        \quad - Sentence Evidence & 89.63 & 89.77\\
        \quad - Fragment Evidence & 89.54 & 89.43\\
        \quad - Phrase Evidence & 89.76 & 90.01\\\hline
        Baseline (ELECTRA$_{base}$)  & 70.20 & 69.28\\
        Mugen on ELECTRA$_{base}$ & 72.01 & 72.56\\\hline
	\end{tabular}
}
\end{table}

To make a brief analysis of the extracted evidence in three different granularities, we execute a series of ablation studies to fix each integrating weight coefficient of evidence to $0$, retaining other coefficients learnable.
Results in Table \ref{ablation_result} suggest that, Fragment Evidence as the \emph{middle-grained evidence} places the most important role among all-grained evidence, while Phrase Evidence has the minimum efficiency.
Further, the quantitative tendency is the same from $base$ to $xxlarge$ model magnitude.

Besides, due to the \emph{Sentence Evidence Extractor} in \emph{Mugen} relies on the pre-trained contextual encoder with additional computational cost, we further design an \emph{Ensemble Baseline} to explore the source of gains from \emph{Mugen}.
\emph{Ensemble Baseline} combines the $[CLS]$ embedding vectors of the baseline and the contextual encoder in \emph{Sentence Evidence Extractor}.
In detail, the two above embeddings are spliced into an integrated embedding in the size of $\mathbb{R}^{2H}$, and a linear feedforward layer is employed, to reduce the dimension of the integrated embedding to $\mathbb{R}^{H}$.
In general, this baseline can be regarded as an enhanced baseline with almost \textit{all additional parameters and pre-trained data} in \emph{Evidence Extractor}.

As shown in Table \ref{ablation_result}, compared to the improvements of the other experimental models, the actual gains from additional neural architectures and pre-trained data in \emph{Ensemble Baseline} are marginal.
It indicates that, most performance gains of \emph{Mugen} are from the extraction and integration of multi-grained evidence.

We also implement \emph{Mugen} based on other encoder baselines, and achieve significant improvements.
The performance of \emph{Mugen} implemented on ELECTRA \cite{clark2019electra} is shown in Table \ref{ablation_result}.
The consistent and significant improvements over various baselines verify the universal effectiveness of \emph{Mugen}.

\subsection{The Roles of Multi-grained Evidence}
\begin{figure}[ht]
\centering
\includegraphics[scale=0.45]{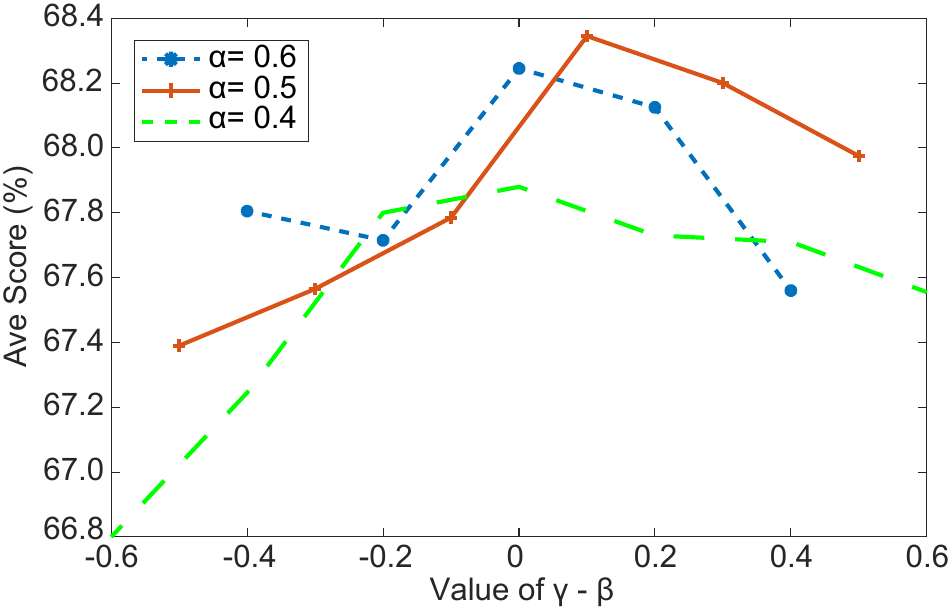}
\caption{The performance curves of evidence with different weight coefficients. For example, point ($0.2$, $68.13$) on line ``$\alpha=0.6$” means when we set $\alpha=0.6, \beta=0.1, \gamma=0.3, \sigma=0$, the average score of \emph{Mugen} on development and test sets of DREAM is $68.13$.}
\label{roles}
\end{figure}
To make a comprehensive analysis of multi-grained evidence, we set and adjust weight coefficients manually and draw the performance curves in Figure \ref{roles}.
To make the figure more intuitive, we set $\sigma=0$ to mask the Phrase Evidence due to its minor contribution, and a specialized analysis of Phrase Evidence will be given later.

The figure depicts that, by allocating a little more proportion to Fragment Evidence than Sentence Evidence, \emph{Mugen} can achieve the best performance.
It indicates Fragment Evidence plays a dominant role to provide guidance information, as well as Sentence Evidence plays an important supporting role, which is consistent with the results in ablation studies.
This finding reveals a reading strategy that, one should refer to the surrounding context (sentencess) to get a comprehensive explanation and to know how to utilize critical evidence fragments.

In addition, paying attention to multi-grained evidence and the original passage in a balanced way ($50\%$ v.s. $50\%$) seems to lead to better performance.
The original passage provides a protective measure to reduce the negative impact of inaccurate evidence extraction, and that is one reason we retain the original passage for information integration in \emph{Mugen}.

To study whether \emph{fine-grained evidence} in phrase level deserves more attention, we fix $\sigma$ to $0.1$ and retain $\alpha, \beta, \gamma$ learnable, leading to a $0.39\%$ drop in average score.
It indicates that, models should not overly depend on \emph{fine-grained evidence}, because evidence at the phrase level may be spliced directly\footnote{We also try to splice them with some punctuation like ``,”, but it does not matter.} and lack complete linguistic structure.

However, combined with the positive effect in ablation studies, \emph{fine-grained evidence} can deliver some detailed information in the form of \textbf{``holes”} just like the example in Figure \ref{example}.
A continuous evidence fragment may bring noisy information to the model like ``\emph{... each focusing on specific topics teaching Hoosiers how to ...}” in the given example, since there exists critical information located at its front and back.
With \emph{fine-grained evidence}, \emph{Mugen} can extract the critical information effectively and dig out the useless information in \emph{middle-grained evidence} in the \textbf{``holes”}.

Finally, to evaluate whether the above analysis is consistent with the characteristics of the multi-choice MRC datasets, we randomly extract $100$ evidence-requiring cases in RACE, DREAM and CosmosQA respectively.
According to the evidence type that provides the most comprehensive information with the least redundant text, we find \textit{coarse-}, \textit{middle-} and \textit{fine-grained} evidence accounts for $27\%/54\%/19\%$ in RACE, $24\%/58\%/18\%$ in DREAM, and $32\%/55\%/13\%$ in Cosmos QA.
The evidence-type distributions are consistent with the above conclusions, showing the effectiveness of the extraction of finer-grained evidence and the integration of multi-grained evidence, as well as justifying the design of the proposed model.

\subsection{Integration and Interaction of Evidence}
The aforementioned experimental results show that, for multi-choice MRC tasks, models can obtain statistically significant performance gains from the simple integration of multi-grained evidence.
Based on the above conclusion, a further question is that, whether the performance gains can be further amplified by elaborate designs that focus on the features of multi-grained evidence.

According to previous researches, typical information enhancement methods mainly include the design of integration strategies \cite{wang2018open, zhao2022poinet} and the modeling of interaction mechanisms \cite{zhang2020dcmn, zhu2020dual}.
For the multi-grained evidence enhancement in this work, the design of integration strategies aims to make evidence in each granularity have the most appropriate contribution to the model prediction;
while evidence interaction mechanisms utilize special neural networks to enrich evidence embedding vectors by the fusion of the evidence in other granularities.
In this section, we implement several \textbf{integration strategies} and \textbf{interaction mechanisms} for multi-grained evidence, to explore possible further gains as well as determine the most effective designs for \emph{Mugen}.

1) \textbf{Voting Integration Strategy}.

In this strategy, four embeddings pass the classifier respectively and \emph{Mugen} uses a majority vote of their predictions.
Based on weight coefficients of evidence, this strategy can be divided into equal voting and weighted voting:
$$\mathcal{L}_{equal}=\sum CELoss(O_i,O_{gold}),$$
$$\mathcal{L}_{weighted}=\sum \theta_i \times CELoss(O_i,O_{gold}),$$
where $i\in\{pas$, $sen$, $fra$, $phr\}$, $\theta_i$ is a learnable weight coefficient, and $O_i$ is the predicted option.

2) \textbf{BiGRU Interaction Mechanism}.

In MRC field, numerous works utilize GRU (Gate Recurrent Unit) or BiGRU to obtain enhanced contextual representation \cite{zhong2019coarse,zhang2020dcmn}.
Inspired by the above studies, we employ a series of BiGRU modules to execute the interaction of the evidence in each granularity:
$$\overrightarrow{h_{phr}} = GRU(e_{phr}, 0),\; \overrightarrow{h_{fra}} = GRU(e_{fra}, \overrightarrow{h_{phr}}),$$
$$\overrightarrow{h_{sen}} = GRU(e_{sen}, \overrightarrow{h_{fra}}),\; \overrightarrow{h_{pas}} = GRU(e_{pas}, \overrightarrow{h_{sen}}),$$
where $h$ is the last hidden representation of GRU.
Take Phrase Evidence as an example, $\overleftarrow{h_{phr}}$ can be obtained similarly, and the interacted representation $E_{phr}$ is generated as:
$$E_{phr}=feedforward(\overrightarrow{h_{phr}} \oplus \overleftarrow{h_{phr}}) \in \mathbb{R}^H,$$
where $\oplus$ is the vector connection operation.
In this mechanism, the original evidence representations will be replaced by above interacted representations for integration.

3) \textbf{Attention Interaction Mechanism}.

We also employ attention-based modules to produce more precise interacted representations.
Take Phrase Evidence as an example, the calculation process is shown as:
$$Att(e_{phr}, e_{*})=softmax(\frac{e_{phr} \cdot e_*^T}{\sqrt{d_k}})e_*,$$
$$E_{phr}=\oplus\{Att(e_{phr}, e_{*})\}W_{phr},$$
where $* \in \{fra, sen, pas\}$, $d_k$ denotes the dimension of \emph{Key} vector, $\oplus\{\}$ denotes the vector connection operation and $W_{phr}$ is a learnable matrix.

\begin{table}[ht]
\caption{\label{strategies} Studies of evidence integration strategies and interaction mechanisms of \emph{Mugen} on DREAM, reported by accuracy ($\%$).}
\centering{
	\begin{tabular}{l|c|c}
	    \hline	\bf Model & \bf Dev & \bf Test  \\
		\hline \hline
        Baseline (ALBERT$_{base}$)  & 65.74 & 65.56\\
        Mugen & 68.83 & 68.69\\\hline
        \quad + Equal Voting Strategy & 66.98 & 66.67\\
        \quad + Weighted Voting Strategy & 67.55 & 67.41\\\hline
        \quad + BiGRU Interaction Mechanism & 68.50 & 68.25\\
        \quad + Attention Interaction Mechanism & 69.07 & 68.83\\\hline
	\end{tabular}
}
\end{table}

Results of various evidence integration strategies and interaction mechanisms of \emph{Mugen} are shown in Table \ref{strategies}, which illustrate that:

1) Among the above \textit{evidence integration strategies}, the direct embedding integration strategy in the original \textit{Mugen} is better than the voting strategy, regardless of loss function types.
Answer prediction relying on only single-grained evidence is inaccurate\footnote{For example, ALBERT$_{base}$ with only Fragment Evidence gets an average score of $58.98$ on DREAM.}, and vote strategy may be heavily hindered by low-accurate models.

2) Among the above \textit{evidence interaction mechanisms}, to our surprise, \textbf{BiGRU Interaction Mechanism} performs worse than the original \textit{Mugen}, which has no interaction mechanism.
It indicates that, improper evidence interaction mechanisms may bring negative impacts on the model.
On the contrary, despite \textbf{Attention Interaction Mechanism} brings marginal improvement, the increase of parameters causes disproportionate computational cost like \cite{zhu2020dual}.

According to the above empirical studies, we conclude that: compared to the original \emph{Mugen}, the further gains by the proposed integration strategies and interaction mechanisms are marginal.
Thus, we retain the original design of \emph{Mugen}, due to its lite scale and adequate improvement.




\subsection{Studies on the Evidence Extractor}
As we state in Section III.C, benefit from the pre-trained encoders, \emph{Evidence Extractor} ensures the quality of the segmentation and extraction of Fragment and Phrase Evidence.
In this section, we attempt to explore the sensitivity of \emph{Mugen} to the \emph{Evidence Extractor}, where the extractor lacks sufficient pre-training or fine-tuning, and the extracted evidence is at a relatively low quality.
In detail, we design three comparative baselines to study the sensitivity to \emph{Sentence Evidence Extractor}, one baseline to study the \emph{Phrase Evidence Extractor}, and three other baselines to explore the accuracy of the design of the proposed multi-grained evidence:

1) \textbf{Weakened Mugen}.

In this baseline, we tune the contextual encoder of \textit{Sentence Evidence Extractor} on SQuAD 2.0 with only one training epoch to make the fine-tuning process inadequate\footnote{As a result, the performance of Exact Match (EM) on SQuAD 2.0 drops from $79.21$ ($2$ training epochs, for the original contextual encoder) to $73.17$ ($1$ training epoch).}, keeping other processes and settings unchanged, to make the fine-tuning process inadequate.

2) \textbf{Attention Mugen}.

\begin{figure}[ht]
\centering
\includegraphics[scale=0.85]{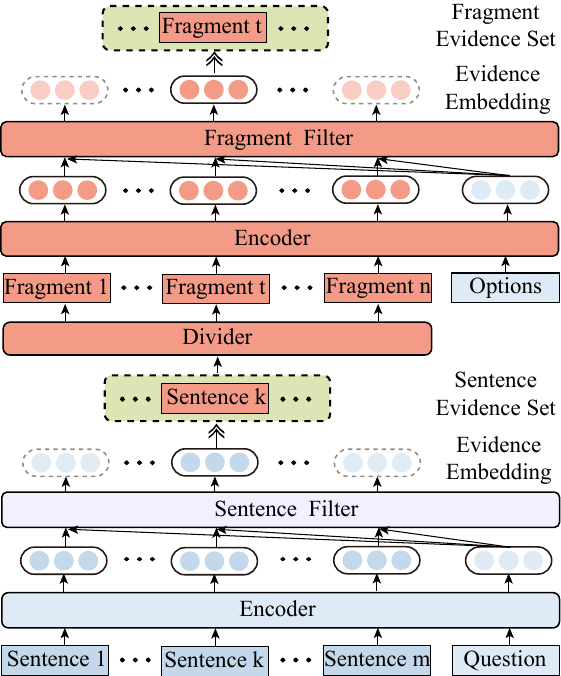}
\caption{The structure of \emph{Sentence Evidence Extractor} in \emph{Attention Mugen}, without pre-trained encoder to extract evidence.}
\label{sentence_extractor_fig}
\end{figure}

In this baseline, we remove the fine-tuning process of the encoder in \emph{Sentence Evidence Extractor}, by using attention calculation for evidence extraction.
Figure \ref{sentence_extractor_fig} shows the architecture of its \textit{Sentence Evidence Extractor}.

\begin{table*}[ht]
	\caption{\label{transfer} Transfer results on the development set of \textit{Mugen} and its baseline, reported by accuracy ($\%$).}
	\centering{
		\begin{tabular}{l|c|c|c|c}
		    \hline	\bf Model & \bf DREAM & \bf Cosmos QA & \bf MCTest 160 & \bf MCTest 500\\
			\hline \hline
            ALBERT$_{base}$ & 65.74 & 63.12 & 73.88 & 76.88\\
            Mugen on ALBERT$_{base}$ & 68.83 & 65.61 & 82.02 & 84.71\\\hline
            Transferred ALBERT$_{base}$ & 66.42 & 43.88 & 81.11 & 82.33\\
            Transferred Mugen on ALBERT$_{base}$ & 69.26 & 48.60 & 87.78 & 88.00\\\hline
		\end{tabular}
	}
\vspace{-0.3cm}
\end{table*}

In general, the above \textit{Sentence Evidence Extractor} is similar to the \textit{Phrase Evidence Extractor} in the original \textit{Mugen}, and the \textit{Encoder} is the same as the one in \textit{Phrase Evidence Extractor}.
In detail, the \emph{Sentence Filter} in \emph{Attention Mugen} computes sentence correlation scores for the embeddings of all input sentences and the given question;
while the \emph{Fragment Filter} computes fragment correlation scores for the embeddings of all input fragments and spliced options, and retains the fragment with the highest score\footnote{In this module, we set the maximum element number to $1$ for Sentence Evidence and Fragment Evidence, which is the same in most cases of the original \textit{Mugen}.}.
To divide the extracted sentences into fragments, we analyze extracted fragments in the original \emph{Mugen}, finding a large proportion of them are clauses divided by pause punctuation (like ``,”, ``-” in the example in Figure \ref{example}).
Thus, the \emph{Divider} in this module divides the sentence into fragments based on all pause punctuation.

3) \textbf{TF-IDF Mugen}.

In this baseline, we remove the encoder in \emph{Sentence Evidence Extractor} directly, where a heuristic TF-IDF method is employed to extract Fragment Evidence with similarity calculation of the context and question-option pairs.
The process of \emph{TF-IDF Mugen} is similar to \emph{Attention Mugen}, except that \emph{TF-IDF Mugen} calculates similarity scores by the TF-IDF method \cite{salton1988tfidf} instead of dot production of vectors.

4) \textbf{TF-IDF Phrase Mugen}.

Referring to \emph{TF-IDF Mugen}, this baseline removes the encoder in \emph{Phrase Evidence Extractor} and provides low-quality Phrase Evidence with the TF-IDF similarity calculation method, but its \emph{Sentence Evidence Extractor} is the same as the original one.

5) \textbf{Sliding Window Mugen}.

We make this design to explore the impact of the accuracy of multi-grained evidence on model performance.
In this design, we employ sliding windows to extract multi-grained evidence with fixed lengths.
The computational method for length-fixed evidence is similar to the \textit{Phrase Evidence Extractor} in the original \textit{Mugen}, where a length-fixed sliding window traverses the entire valid contextual text, and calculates the correlation score of each extracted text segment and the question-option pair in turn, according to the formula we state in Section III.C.
Ultimately, the text segment with the highest correlation score is defined as the extracted evidence\footnote{When the length of the traversable text is less than the sliding window, the entire traversable text is defined as the evidence.}.
For the lengths of sliding windows, we design two different combinations:

\begin{itemize}
    \item \textbf{Tri-Gram-Bi-Gram-Word}: the lengths of the three sliding windows are respectively $3$, $2$ and $1$ word(s);
    \item \textbf{Average Length}: lengths of the three sliding windows are the respective average lengths of the multi-grained evidence in the original \textit{Mugen}. For the DREAM dataset, the lengths of sliding windows are $11$, $6$ and $4$ respectively.
\end{itemize}

6) \textbf{Damaged Mugen}.

To study the accuracy of multi-grained evidence, in this baseline, the original extracted evidence at each granularity is randomly damaged by adding or deleting several words ($1$ for Phrase Evidence and $2$ for others) on its front and back textual boundaries.
We additionally set the evidence at each granularity to have at least one word to avoid excessive damage, and the operations beyond the valid contextual text are filtered.

\begin{table}[ht]
\caption{\label{improvement_result} Comparative experiment results on DREAM, reported by accuracy ($\%$).}
\centering{
	\begin{tabular}{l|c|c}
	    \hline	\bf Model & \bf Dev & \bf Test  \\
		\hline \hline
        Baseline (ALBERT$_{base}$) & 65.74 & 65.56\\
        Ensemble Baseline & 66.87 & 66.73\\
        Mugen & 68.83 & 68.69\\\hdashline
        Weakened Mugen & 68.23 & 68.15\\
        Attention Mugen & 67.84 & 67.55\\
        TF-IDF Mugen & 67.59 & 67.21\\
        TF-IDF Phrase Mugen & 68.48 & 68.40\\
        Sliding Window Mugen (Tri-Gram-Bi-Gram-Word) & 66.47 & 66.29\\
        Sliding Window Mugen (Average Length) & 67.70 & 67.61\\
        Damaged Mugen & 67.65 & 67.47\\\hline
	\end{tabular}
}
\vspace{-0.3cm}
\end{table}

We evaluate these baselines on DREAM based on ALBERT$_{base}$, as Table \ref{improvement_result} shows.
The results indicate three main conclusions:

1) The performances of the experimental baselines with low-quality evidence are still significantly better than \textit{Ensemble Baseline} (stated in Section V.A), which further proves the performance gains of \emph{Mugen} are mainly from the integration of multi-grained evidence;

2) The proposed \emph{Mugen} has satisfactory robustness to the quality of evidence (or the design of \emph{Evidence Extractor}), and the performance of \emph{Mugen} increases with the more powerful encoding ability of its \emph{Evidence Extractor} and more accurate multi-grained evidence.

3) Length-fixed textual evidence without grammatical structure and semantic integrity brings significant damage to the model performance, which proves the gains of \emph{Mugen} are mainly from the accurate and linguistic multi-grained evidence design.

\subsection{Transferability Studies}
To further verify the generalizability and robustness of \emph{Mugen}, we implement a series of transfer experiments.
We train \emph{Mugen} and its baseline on RACE and evaluate them on DREAM, Cosmos QA and MCTest respectively, obtaining transfer results on the development set as Table \ref{transfer} shows.

In Table \ref{transfer}, transferred \textit{Mugen} obtains more consistent performance improvements than its baseline on various out-of-domain datasets, proving the generalizability and robustness of \emph{Mugen}.
Furthermore, transferred \emph{Mugen} even performs better than the original in-domain trained \emph{Mugen} on DREAM and MCTest, indicating models may benefit from larger out-of-domain training datasets like RACE.
On the contrary, transfer results on Cosmos QA drop significantly, due to its disparate data collection sources and question-type proportion.

\subsection{Error Case Analysis}
\begin{figure*}[ht]
\centering
\includegraphics[scale=0.44]{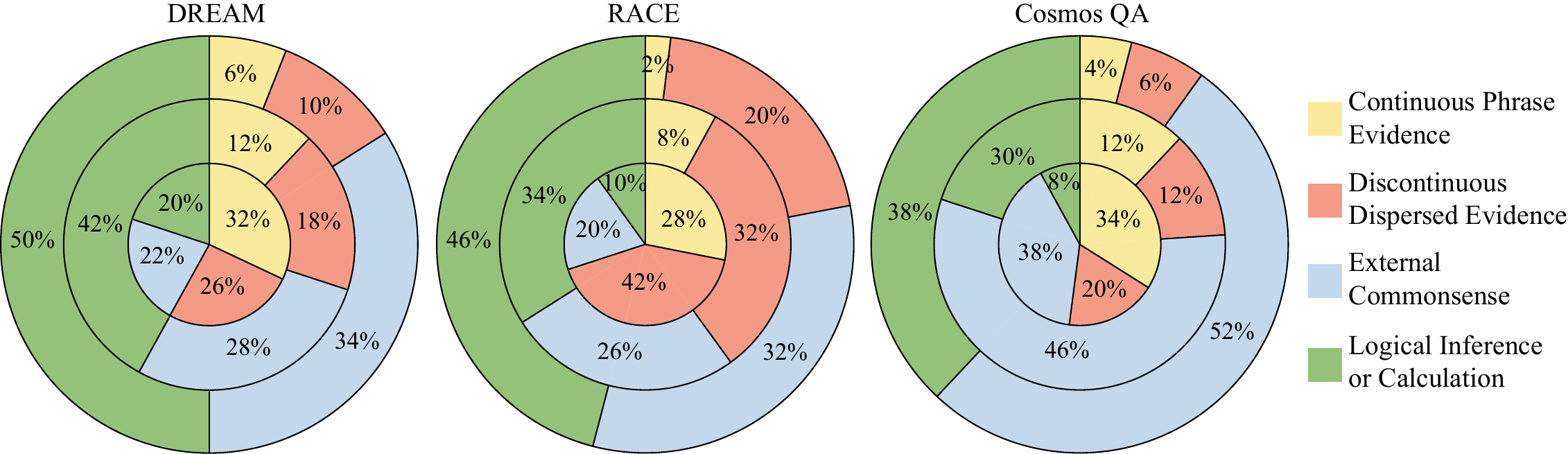}
\caption{Error cases on DREAM, RACE and Cosmos QA. For each chart, parts from the innermost circle to the outermost circle represent the cases in: 1) the original dataset; 2) error cases predicted by ALBERT$_{xxlarge}$; and 3) error cases predicted by \emph{Mugen} on ALBERT$_{xxlarge}$.}
\label{error_cases}
\vspace{-0.3cm}
\end{figure*}

To explore potential further improvement, on DREAM, RACE and Cosmos QA three datasets, we randomly extract $50$ examples respectively in 1) the original dataset; 2) error cases predicted by ALBERT$_{xxlarge}$; and 3) error cases predicted by \emph{Mugen} on ALBERT$_{xxlarge}$\footnote{We do not perform the above operations on MCTest due to its minor dataset scale.}.
We divide them into different types according to ``the most decisive information for correct prediction” and draw the analysis donut chart as Figure \ref{error_cases}.

The above chart depicts that \emph{Mugen} has an excellent ability to solve questions requiring evidence integration.
Take RACE as an example, compared to its baseline model, \emph{Mugen} benefits from the \textit{middle-} and \emph{fine-grained evidence}, and the proportion of the error cases requiring continuous phrase evidence receives an additional $6\%$ reduction.
In the same way, the integration of \emph{multi-grained evidence} helps to solve an additional $12\%$ of the cases requiring discontinuous dispersed evidence, like the example in Figure \ref{example}, as well as the following conversation in Table \ref{conversation}.

\begin{table}[ht]
\centering
\caption{\label{conversation} A sample conversation requiring inference from multi-grained evidence from DREAM.}
\begin{tabular}{|p{8cm}|}
\hline \bf Conversation 2\\ \hline
\emph{...}\\
\emph{A: And are there other materials I would need to send in addition to the application form?}\\
\emph{B: Uh, yes. You would need to send in \uline{a \$35 \textbf{non-refundable application fee} [Uh-huh], a \textbf{sponsorship form indicating} who will be responsible financially for the student while studying in our program, and a \textbf{bank statement}} showing that you or your sponsor has sufficient funds to cover tuition expenses and living costs for the entire year of study.}\\
\emph{...}\\\hline
Q: \emph{Which one was NOT mentioned as part of the application packet a student must send to the center?}\\
A. \emph{sponsorship form}\\
B. \emph{high school transcripts} (golden answer)\\
C. \emph{application fee}\\\hline
\end{tabular}
\end{table}

In detail, the underline context is the extracted Fragment Evidence, the sentence containing it is the Sentence Evidence, and the Phrase Evidence is marked in bold.
As a typical instance of ``discontinuous dispersed evidence for answer prediction” in Figure \ref{error_cases}, the integration of multi-grained evidence helps to infer out ``\textit{the item not mentioned}”, while relying on one single-grained evidence may not predict the golden answer ultimately (\emph{coarse-grained evidence} may overemphasize interference information while \emph{fine-grained evidence} may lack contextual explanation).

This chart also reveals the challenging cases for \emph{Mugen}.
Take RACE as an example, compared to its baseline, the proportion of error cases caused by the lack of external commonsense increases by $6\%$, indicating \emph{Mugen} can benefit from explicit commonsense injecting.
Another challenge is the questions requiring logical inference or calculation, where the typical types are attribute sorting, location description and numerical calculation, instead of the inference based on evidence chains.

Furthermore, the statistics of error cases on DREAM and Cosmos QA has a similar tendency to RACE, emphasizing the effectiveness of \textit{Mugen} to questions with continuous or discontinuous evidence.
Surprisingly, in Cosmos QA, the proportion of questions requiring external commonsense does not increase as significantly as RACE and DREAM, indicating the commonsense in Cosmos QA may be inferred by multi-grained evidence integration within the passages.

\begin{figure}[ht]
\centering
\includegraphics[scale=0.42]{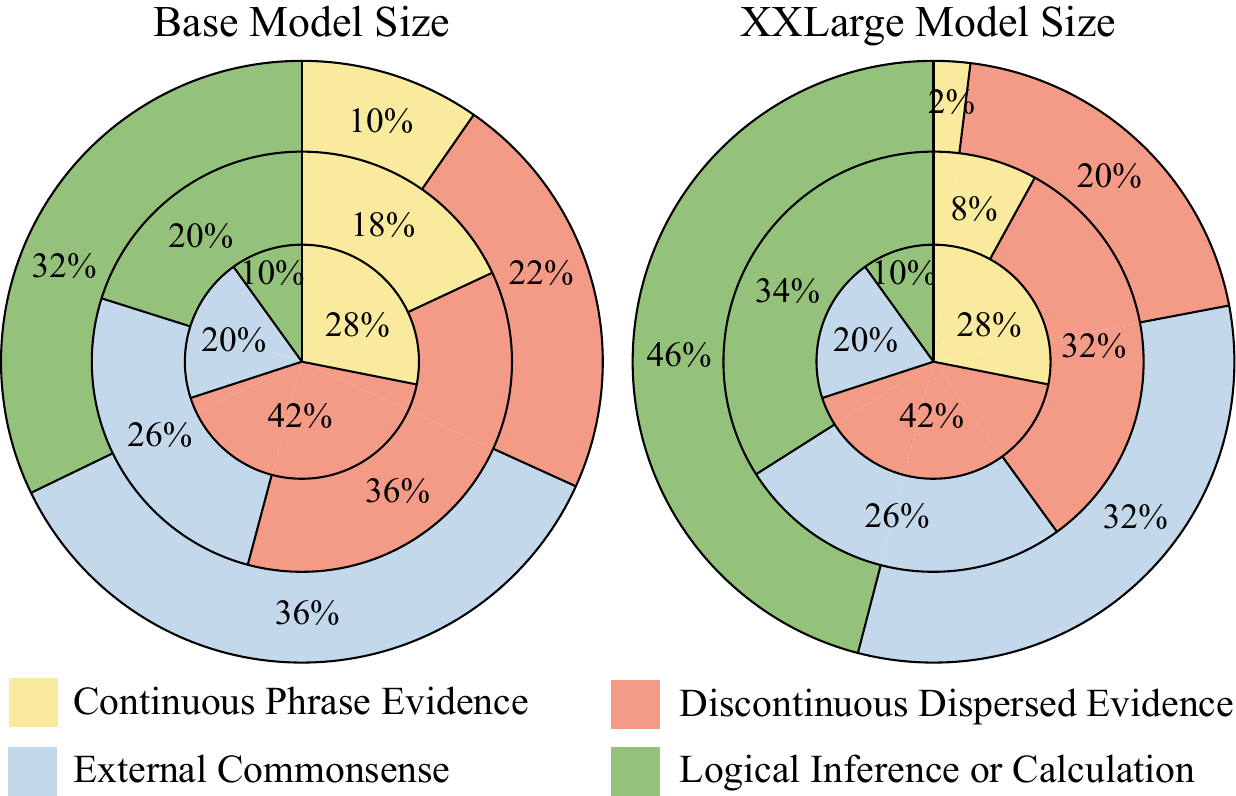}
\caption{The type distribution of error cases predicted by models in different sizes on RACE. For each chart, parts from the innermost circle to the outermost circle represent the cases in: 1) the original dataset; 2) error cases predicted by ALBERT; and 3) error cases predicted by \emph{Mugen}.}
\label{race_error_cases}
\vspace{-0.3cm}
\end{figure}

In addition, we also make statistics on the error-type distribution of \emph{Mugen} in \textit{base} and \textit{xxlarge} two parameter magnitudes.
As Figure \ref{race_error_cases} and Table \ref{race_result} show, compared to \emph{Mugen} on ALBERT$_{xxlarge}$, \emph{Mugen} on ALBERT$_{base}$ reduces the evidence-requiring error cases with a larger proportion and obtains more significant performance improvement.
One main reason is, due to the limited parameter magnitude, the baseline model ALBERT$_{base}$ does not have strong abilities of text encoding and information integration like ALBERT$_{xxlarge}$, and can gain more benefits from the multi-grained evidence integration in \emph{Mugen}.

\section{Conclusion}
In this work, we propose a general-purpose model enhancement design that integrates multi-grained evidence comprehensively, called \emph{\textbf{Mu}lti-\textbf{g}rained \textbf{e}vidence i\textbf{n}ferencer (\textbf{Mugen})}, to make up for the inability to deliver evidence in different granularities in existing studies.
With integration and inference, \emph{Mugen} achieves substantial improvement on four multi-choice MRC benchmarks: RACE, DREAM, Cosmos QA and MCTest with all passing significance tests, which indicates the superiority of multi-grained evidence integration and points out a promising research direction.

\ifCLASSOPTIONcaptionsoff
  \newpage
\fi



%


\bibliographystyle{IEEEtran}
\bibliography{mugen}
\end{document}